# Incipient Slip-Based Rotation Measurement via Visuotactile Sensing During In-Hand Object Pivoting

Mingxuan Li[1], *Graduate Student Member*, *IEEE*, Yen Hang Zhou[1], Tiemin Li[1], and Yao Jiang[1], *Member*, *IEEE*

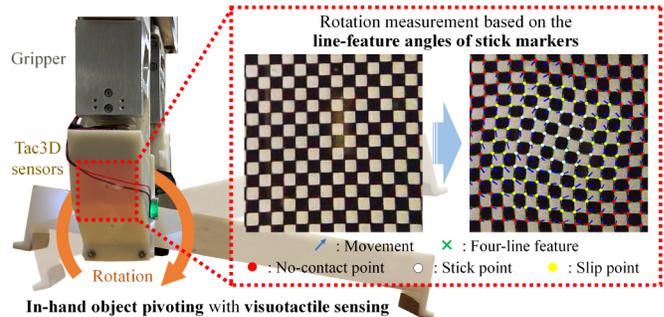

Fig. 1. Online rotation measurement based on the line-feature angles of stick marker points during pivoting.

*Abstract*—In typical in-hand manipulation tasks represented by object pivoting, the real-time perception of rotational slippage has been proven beneficial for improving the dexterity and stability of robotic hands. An effective strategy is to obtain the contact properties for measuring rotation angle through visuotactile sensing. However, existing methods for rotation estimation did not consider the impact of the incipient slip during the pivoting process, which introduces measurement errors and makes it hard to determine the boundary between stable contact and macro slip. This paper describes a generalized 2-d contact model under pivoting, and proposes a rotation measurement method based on the line-features in the stick region. The proposed method was applied to the Tac3D vision-based tactile sensors using continuous marker patterns. Experiments show that the rotation measurement system could achieve an average static measurement error of 0.17°±0.15° and an average dynamic measurement error of 1.34°±0.48°. Besides, the proposed method requires no training data and can achieve real-time sensing during the in-hand object pivoting.

## I. INTRODUCTION

In robotic manipulation tasks, extrinsic dexterity has been proven to compensate for the shortcomings of underactuated grippers, thereby effectively improving the grasping and manipulation capabilities [1]. Among them, the problem of rotating the grabbed object without regrasping is called object pivoting [2]-[4], which is achieved by adjusting the grasp force to control rotation affected by gravity [5] or driven under gravity [6]. Such tasks usually require re-locating the in-hand object to a specific rotation angle. Besides, pivoting is also a typical grasp failure event. When the grasping position is far from the object's center of gravity, the large torque at the contact point may cause the object to rotate. Thus, the rotation angle perception during object pivoting is of great significance for improving the dexterity and stability of robotic hands.

In traditional studies, robotic manipulation tasks usually rely on vision to detect the contact position and pose [7]. However, the effects of external illumination and the shading from the gripper may lead to inaccuracy in measurement. Also, the object's physical properties and the contact states cannot be reflected. In contrast, tactile sensors can perceive such information and benefit the construction of an effective closed-loop control framework [3], [8]. Among existing tactile sensing technologies, visuotactile sensing (also known as vision-based tactile sensing) [9] is a promising solution. The visuotactile sensor uses a built-in camera to capture the contact deformation of a soft elastomer and reconstruct the contact characteristics using either the marker displacement method [10] or the photometric stereo method [11]. Such sensors were proved suitable for measuring pivotal rotation [12], [13]. However, there is an individual state during the object pivoting that is rarely discussed: the incipient slip [14]. In practice, detecting only macro rotational slip can lead to measurement errors and the overly optimistic estimate of stable grasping. In addition, existing methods are facing difficulties in handling special cases like translation, small contact areas, and special contact shapes [12], [13]. The above issues require research on contact mechanics during pivoting.

This paper describes a contact model for elastomer-based pivotal rotation computation, and presents a method based on incipient slip detection for measuring the rotation angle during pivoting. The method was applied to the CMP-based Tac3D sensor [15], and has achieved effective application of online rotation angle measurement on the robot system [see Fig.1]. Experiments showed that the described method could be applied to objects with unknown geometries and physical properties. It achieved a static measurement accuracy of 0.17°±0.15° and a dynamic accuracy of 1.34°±0.48° without any prior information. The core advantages are as follows:

- High precision and accuracy; less affected by contact shape, contact area, and translational displacement.
- Clear physical meaning; no training dataset required.

## II. RELATED WORK

### A. Object Rotation Measurement

In the current research, in-hand object rotation has been regarded as a type of slippage (another one is translation).

*This work was supported by the National Natural Science Foundation of China under Grant 52375017 and the iCenter Star of Innovation Program from the Fundamental Industry Training Center, Tsinghua University.

[1]Mingxuan Li, Yen Hang Zhou, Tiemin Li, and Yao Jiang (corresponding author) are with the Institute of Manufacturing Engineering, Department of Mechanical Engineering, Tsinghua University, Beijing 100084, China (e-mail: mingxuan-li@foxmail.com; zhouyanh23@mails.tsinghua.edu.cn; litm@mail.tsinghua.edu.cn; jiangyao@mail.tsinghua.edu.cn).

Yamada *et al.* used a stress-rate sensor to achieve dynamic rotational slip detection measurement [16]. Studies based on different neural networks could detect the rotation angle by learning tactile features [6], [17], [18]. In relevant studies using visuotactile sensors, Zhang *et al.* built an LSTM-based neural network to detect the contact events, including rotation [19]; Kolamuri *et al.* formulated the least-squares solution to measure the rotation [12]; Castaño-Amorós *et al.* proposed a system that used the tactile segmentation neural network to segment the contact region and estimate the rotation [13].

Although the above works aimed to detect object rotation, the physical properties they actually measured are different. Some studies focused on the relative rotation of objects to the contact surface, which should be considered as the macro slip behaviors [6], [13], [16]-[18]. Other studies concentrated on the rotation of the contact surface itself [12], [19], while it is equivalent to assuming fully sticking. However, in the rotation process before grasping failure, the slip and stick states usually coexist on the contact surface (i.e., incipient slip). According to Dong *et al.* [20], the incipient slip during rotation could cause a deviation between the ground truth and the estimated value. These issues inspire us to start with the research of incipient slip states under object pivoting.

*B. Incipient Slip Detection*

Incipient slip has been considered as a state between after stable contact and before macro slip. During the contact, the contact region can be divided into the stick and slip regions from the inside out. As the shear and torsion increases, the slip region gradually expands while the stick region shrinks. When the stick region disappears completely, the macro slip state reaches. Early studies [21], [22] focused on predicting the onset of macro slip, but were unable to quantify the incipient slip. Some evaluation metrics using normal and shear stresses were then proposed [23], [24]. Due to the hardware limitation, such methods did not achieve enough performance.

In recent studies, visuotactile sensors were proved suitable for fine quantitative measurements of incipient slip. Ito *et al.* estimated the stick region by the movement of markers [25]; Yuan *et al.* described the inhomogeneity of the displacement field based on entropy and proved that this metric was related to the incipient slip [26]; Dong *et al.* compared the marker displacement at the contact region edge to the maximum displacement to determine the stick ratio [27]. These methods only considered rigid contact objects. Experiments in [20] proved that rigidity assumption-based methods perform poorly when dealing with deformable objects. Sui *et al.* proposed an incipient slip model for soft objects, which can apply to more household objects [28]. Based on Sui's work, we propose a method for incipient slip and rotation angle detection using the angular change of line features.

III. THEORY

This section introduces examples and mechanical models of incipient slip under pivoting, and proposes a theoretical model of using line features to determine the stick/slip state and estimate the rotation angle.

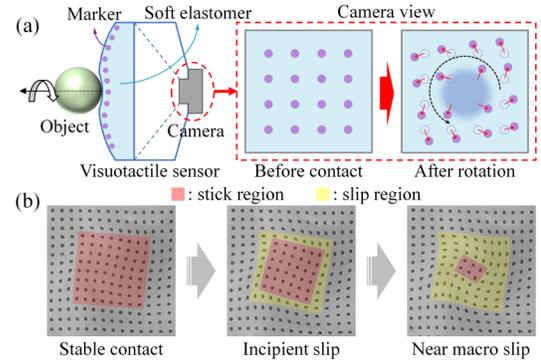

**Fig. 2.** (a) Principle and structure of visuotactile sensors. (b) The incipient slip phenomenon when a cube-shaped object contacts the soft elastomer.

*A. Incipient Slip during Pivoting*

To explain the incipient slip phenomenon mentioned in Section II-B and its effects, we use the marker pattern-based visuotactile sensor for sample demonstration. As shown in Fig. 2(a), a visuotactile sensor can sense the contact state of in-hand objects by measuring the contact deformation through detecting and tracking the markers. Fig. 2(b) shows the variation of the tactile images obtained when a cube-shaped object is pivoting on the contact surface. At the initial trial, the portion of the contact surface in contact with the object maintains the same rotation mode as the object. This state is known as stick or stable contact. As the rotation angle increases, the markers near the contact region edges exhibit a hysteresis of motion (i.e., relative motion is produced), which is regarded as slippage. During the rotation, the slip region gradually expands from outside to inside until the stick region disappears completely. At this point, the object is considered to have undergone macro slip. It means that the state change from fully stable contact to macro slip is not a leap but a gradual evolutionary process. The closer a position to the outside of the contact region, the more likely it is to slip. This phenomenon is called the incipient slip, which can be explained by the Coulomb friction model in [14].

*B. A Representation Model of Slip and Rotation*

We describe a generalized 2-d contact model between an object and a sensor's soft elastomer under pivotal rotation, as shown in Fig. 3. The coordinate systems $\{O_S\}$ and $\{O_B\}$ are constructed on the contact surfaces of the elastomer and the object, respectively, and are set to be aligned at the initial moment $t_0$. Pick point $p_S$ in $\{O_S\}$ and $p_B$ in $\{O_B\}$. Initially, the position vectors of $p_S$ and $p_B$ denote $\boldsymbol{r}_S^{O_S}(t_0)$ and $\boldsymbol{r}_B^{O_B}(t_0)$, respectively (the superscripts indicate that they are within the $\{O_S\}$ and $\{O_B\}$ systems, respectively), and these two points are in the same position in space. After time $\Delta t$, let the overall displacement and rotation of $\{O_B\}$ with respect to $\{O_S\}$ be $\Delta\boldsymbol{r}_c^{O_S}$ and $\Delta\boldsymbol{R}_c^{O_S}$, respectively (with the rotation angle $\theta$). The position vectors of $p_S$ and $p_B$ are $\boldsymbol{r}_S^{O_S}(t_0 + \Delta t)$ and $\boldsymbol{r}_B^{O_B}(t_0 + \Delta t)$, and the micrometric deformations at that positions denote $\Delta\boldsymbol{r}_S^{O_S}$ and $\Delta\boldsymbol{r}_B^{O_B}$, respectively. By transforming $\Delta\boldsymbol{r}_B^{O_B}$ to the $\{O_S\}$ system, the slip between $p_S$ and $p_B$ can be expressed as

$$\Delta s_p^{O_S} = \Delta r_B^{O_B} + \Delta r_c^{O_S} + \Delta R_c^{O_S} \cdot r_B^{O_B}(t_0 + \Delta t) - \Delta r_S^{O_S}. \quad (1)$$

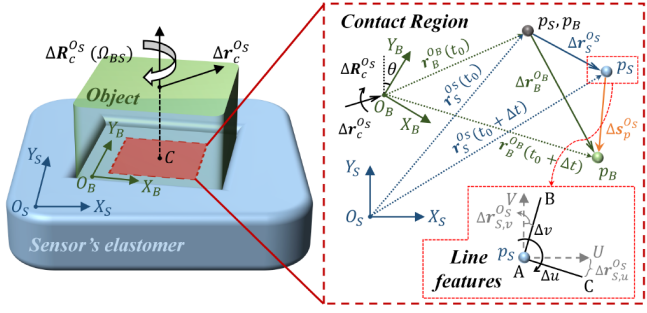

**Fig. 3.** A generalized 2-d contact model between an object and a soft elastomer under pivotal rotation.

According to the derivation in the appendix of [28], the deformation fields of the object and the elastomer under the assumption of an elastic semi-infinite plate approximately satisfy the following relations

$$D_m(\Delta r_{B,n}^{O_B}) = -k \cdot D_m(\Delta r_{S,n}^{O_S}) + \Delta l_{mn}, \quad (2)$$

where $D_m$ denotes the derivative along the m-axis (x or y), $\Delta r_{B,n}^{O_B}$ and $\Delta r_{S,n}^{O_S}$ denote the components of $\Delta r_B^{O_B}$ and $\Delta r_S^{O_S}$ in the n-axis (x or y), $k$ denotes the ratio of the shear elasticity modulus of the soft elastomer to that of the object, and $\Delta l_{mn}$ is related to the contact distribution force and is approximately a constant value on the contact surface.

By selecting the x-axis component in Eq. (1), deriving it for the y-direction (note that the derivative here is taken in $\{O_S\}$), and substituting Eq. (2), we can obtain the equation

$$D_y(\Delta s_{p,x}^{O_S}) = -(k+1) \cdot D_y(\Delta r_{S,x}^{O_S}) + \Delta l_{yx} \\ + (\cos\theta - 1) \cdot \sin\theta + \sin\theta \cdot \cos\theta, \quad (3)$$

where $\theta$ is the rotation angle. Since Sui *et al.* pointed out that Eq. (2) has high consistency with the simulation under small deformation [28], we consider $\Delta t$ is small and introduce small angle approximation in Eq. (3) to obtain the equation

$$D_y(\Delta s_{p,x}^{O_S}) = -(k+1) \cdot D_y(\Delta r_{S,x}^{O_S}) + \Delta l_{yx} + \Omega_{BS}\Delta t, \quad (4)$$

and it can be concluded similarly that

$$D_x(\Delta s_{p,y}^{O_S}) = -(k+1) \cdot D_x(\Delta r_{S,y}^{O_S}) + \Delta l_{xy} - \Omega_{BS}\Delta t, \quad (5)$$

where $\Omega_{BS}$ denotes the angular velocity of $\{O_B\}$ with respect to $\{O_S\}$ during $\Delta t$. Thus, subtracting Eq. (4) from Eq. (5) yields

$$rot(\Delta s_p^{O_S}) = -(k+1) \cdot rot(\Delta r_S^{O_S}) \\ + (\Delta l_{xy} - \Delta l_{yx}) - 2\Omega_{BS}\Delta t. \quad (6)$$

Eq. (6) describes the relationship between the increment of the slip field and the deformation field on the contact surface during pivoting, and *rot* represents the curl. Inspired by Eq. (6), we use line features to reflect curl information, as shown in Fig. 3. We select two orthogonal micro-cluster segments AB and AC with lengths $\Delta u$ and $\Delta v$ through point $p_S$, and establish the coordinate systems UAV. During the time period $\Delta t$, the rotational angular velocity of AB and AC is

$$\omega_{AB} = \lim_{\Delta t \to 0} \frac{\partial(\Delta r_{S,v}^{O_S}/\Delta t)}{\partial u}, \omega_{AC} = -\lim_{\Delta t \to 0} \frac{\partial(\Delta r_{S,u}^{O_S}/\Delta t)}{\partial v}. \quad (7)$$

Considering the impact of angular deformation motion in a 2-d flow field, we select the average value to define the rotational angular velocity of the deformation field at point $p_S$ as

$$\omega_p = \frac{\omega_{AB} + \omega_{AC}}{2} = \frac{rot(\Delta r_S^{O_S}/\Delta t)}{2}. \quad (8)$$

Eq. (8) indicates that the average value of two perpendicular tangents' angular velocity is approximately equal to the curl of the deformation velocity increment at that position. Substituting Eq. (8) into Eq. (6) yields

$$rot(\Delta s_p^{O_S}) = -2(k+1) \cdot \omega_p \Delta t \\ + (\Delta l_{xy} - \Delta l_{yx}) - 2\Omega_{BS}\Delta t. \quad (9)$$

We select two points $i$ and $j$ on the contact surface of sensor's elastomer. According to Eq. (9), it can be calculated that

$$\omega_i - \omega_j = -\frac{rot\left(ds_i/dt - ds_j/dt\right)}{2(k+1)}. \quad (10)$$

By integrating Eq. (10) over time, and considering the uniform continuity of the slip field and its partial derivative function, we can obtain the following relation

$$\Delta\varphi_{ij} = -\frac{rot(\Delta s_{ij})}{2(k+1)}, \quad (11)$$

where $\Delta\varphi_{ij}$ and $\Delta s_{ij}$ denote the differences between $i$ and $j$ of the average rotation angle and the total slip, respectively.

Eq. (11) illustrates that the rotation angle difference of two points on the contact surface can be used to approximate the difference in slip degree (cumulative amount of curl). If a point i in the stick state is determined, it is possible to detect the stick/slip state of j by comparing the rotation angle of their line features. According to Eq. (9), for point i in the stick state, the following relation is always satisfied:

$$\Omega_{BS}\Delta t = -(k+1) \cdot \omega_i \Delta t + \frac{(\Delta l_{xy} - \Delta l_{yx})}{2}. \quad (12)$$

Thus, by integrating Eq. (12) over time, we can obtain

$$\theta = -(k+1) \cdot \varphi_i + \frac{1}{2}(L_{xy} - L_{yx}), \quad (13)$$

where $L_{xy}$ and $L_{yx}$ denote the accumulation of $\Delta l_{xy}$ and $\Delta l_{yx}$. According to [28], when the object stiffness is much larger than that of the elastomer, the values of $k$, $L_{xy}$ and $L_{yx}$ take values close to 0. Therefore, Eq. (13) can be approximated as

$$\theta = -\varphi_i. \quad (14)$$

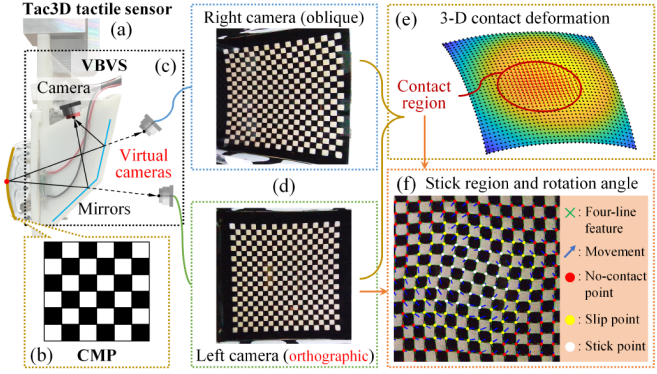

Fig. 4. Rotation angle measurement method. (a) Tac3D tactile sensor. (b) Continuous marker pattern (CMP) [15]. (c) Virtual binocular vision system (VBVS) [31]. (d) Acquired binocular tactile images. The left camera's field of view is set to orthographic. (e) 3-D contact deformation reconstruction and contact area estimation. (f) Stick region and rotation angle measurement.

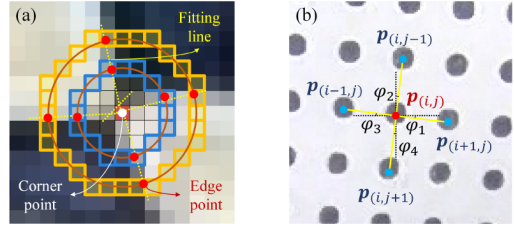

Fig. 5. (a) Calculation of line-feature angles based on a two-layer circular detector. (b) Calculation of line-feature angles when using discrete markers.

Eq. (14) illustrates that the rotation angle of points in the stick region can be approximated to represent the pivotal rotation (the negative sign indicates the opposite normal direction). Besides, Eq. (13) can describe the error caused by the rigid object assumption. The translational displacement of the object due to gravity as the gripper lifts it has been considered in the above derivation. Since the translational terms are not explicitly included in Eqs. (11) and (14), the line feature angles are sensitive to the rotational displacement response and less affected by the translational displacement, which is suitable for rotation estimation. Thus, the process of pivotal rotation measurement includes three steps:

- Find some stick points on the sensor's contact surface;
- Detect the local angle of rotation on the contact surface using available line features to find the whole stick region according to Eq. (11);
- Calculate the average rotation angle of the stick region and estimate the pivoting angle according to Eq. (14).

IV. METHOD

Based on the theory in Section III, we customized a Tac3D sensor and used the approach shown in Fig. 4 to achieve real-time measurement of the pivoting angle at 30 Hz (mainly limited by the built-in camera's performance).

A. Design of Tactile Sensor

Based on the basic structure of Tac3D introduced in Section III-A, two special designs are used to enhance its measurement of rotation. They are not mandatory. Section IV-B describes how visuotactile sensors using monocular view and discrete markers can achieve rotation measurement.

1) Continuous marker pattern (CMP). Our previous work [15] proposed the CMP for enhanced contact information representation. During the pivoting, the CMP can provide highly recognizable edge-line features for angle detection. In addition, the rigid matching property of CMP ensures reliable information transfer under large deformation to tolerate interruptions during the process, thus continuous frame recording is not required. The marker recognition and tracking were done using the feature detection method in [29] and the spin-search algorithm in [30].

2) Virtual binocular vision system (VBVS) containing an orthographic view. A single physical camera's field of view can be divided into left and right parts by cleverly installing multiple mirrors, equivalent to having two virtual cameras for measuring 3-d deformation [31]. Besides, the left view is an orthographic view in order to detect the line features. As a result, both front-view shooting and stereo vision are achieved in Tac3D to ensure compactness and synchronized triggering.

B. Rotation Angle Measurement

The proposed method first determines the contact region using the 3-d deformation field to find the markers in the stick state. The 3-d displacement of markers are obtained by the method in [15]. The largest of normal displacements is used as a threshold to filter the markers with small movement (50% is chosen). A marker $p_0$ close to the contact region's geometric center and has the tangential movement near the average value is assumed to be the stick center. When the deformation vector modulus of $p_0$ exceeds a certain threshold (0.1 mm is chosen), we determine that contact has occurred at that time.

We further distinguish between stick/slip states within the contact region and calculates the stick region's average rotation angle to estimate the pivoting angle. The line feature rotation angles of all markers are detected by the double-layer circular detector introduced in [29] [see Fig. 5(a)]. Since the angular deformation could cause the line features to deviate from vertical, the average angle of the four edge-lines is chosen. Taking the stick center $p_0$ as the starting point, we adopt a strategy based on clustering to segment the stick region. At moment $k$, we let the average angle within the stick region $S_k$ be $\overline{\varphi_k}$, and select a point $p_i$ in the 4-neighborhood of the $S_k$ with the rotation angle $\varphi_i$. According to Eq. (11), $p_i$ is considered sticking when the normalized difference satisfies

$$\Delta\varphi = \frac{|\varphi_i - \overline{\varphi_k}|}{\sqrt{\varphi_i \cdot \overline{\varphi_k}}} < \Delta\varphi_{th}, \quad (15)$$

where $\Delta\varphi_{th}$ is the selected threshold (set to 0.4° in this paper). Therefore, the point $p_i$ is added to $S_k$ and $\overline{\varphi_k}$ is recalculated. The above process is repeated continuously, and $S_k$ keeps growing from the center $p_0$ to the outside until no new 4-adjacent point can be added. According to Eq. (14). The final average angle $\overline{\varphi_k}$ of the stick region $S_k$ is regarded as the angle of pivotal rotation. When the number of markers within $S_k$ is less than 3, we consider that the assumed stick center $p_0$ has slipped and the macro slip state has been reached.

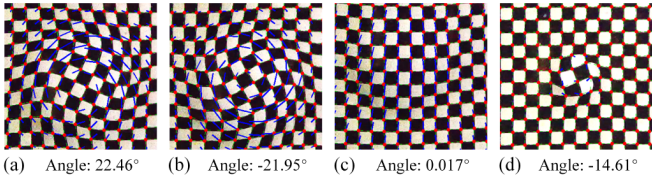

**Fig. 6.** Four types of contact rotation and the measurement results. (a) Clockwise rotation. (b) Counterclockwise rotation. (c) No rotation. (d) Rotation in the small and round contact area.

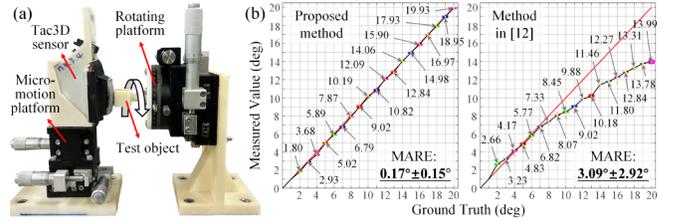

**Fig. 7.** (a) Experimental setup for static measurement evaluation. (b) Evaluation results. The red line denotes the ground truth angle, the black line denotes the measurement angle, and the colored boxes denote the discrete distribution of the repeated test results. MARE denotes the mean absolute rotational error introduced in [13].

Fig. 6 illustrates the results of our method in dealing with different cases. As a result, the proposed method can naturally distinguish between translational and rotational displacements, and suppresses the translation effect while responding sensitively to rotation. For small contact areas (even with round shape), it can work without using the less reliable contour tracking strategies used in [12] and [13].

For visuotactile sensors based on discrete markers, the contact region can be determined based on the translation field dispersion [32] or light subtraction operation [12]. Besides, the line feature calculation can be done as shown in Fig. 5(b). In this approach, the measurement and spatial resolutions are equivalent to the minimum spacing of the marker array.

## V. EXPERIMENTS

In this section, two sets of experiments are performed: the static evaluation of the measurement accuracy and the online measurement in pivoting tasks. In the static experiments, we focus on quantitatively evaluating the nominal accuracy of the proposed method and comparing it with the approach in [12]. In the dynamic experiments, the method's performance of real-time perception in pivoting tasks was mainly considered. To reflect the generalizability, the first experiment's results were not used as a prior reference of calibration for the second experiment.

### A. Static Measurement Evaluation

Fig. 7(a) shows the experimental setup for static evaluation. A standard cube with a side length of 15 mm was chosen as the test object. The Tac3D tactile sensor and the test object were fixed on two three-axis micro-motion platforms, which have a translation accuracy of 0.02 mm. The platform on which the object was mounted was configured with a rotary stage with an accuracy of 0.1°, which could control the object rotation to a defined angle perpendicular to the normal direction of the contact surface. Considering that the signal-to-noise ratio is higher when the rotation angle is small while the macro slip happens when the rotation angle is large, we empirically set the range of the proposed method to 2 to 20 degrees. Within the effective range, we performed 25 to 30 repetitions of the experiment for each integer scale. The methods used could be quantitatively evaluated by comparing the measured angle with the ground truth angle provided by the rotating platform.

Since other relevant methods have used learning-based techniques [13], [19] or did not show details [20], [33], we only made objective comparisons with the method proposed by Kolamuri *et al.* [12] [see Fig. 7(b)]. Here, our own implementation version was used since their code was not available. The two approaches possessed high precision since their data dispersion was small. However, the method in [12] exhibited a tendency of a gradual increase in the average error as the rotation angle increased. It is in line with the explanation of experiments in [12]: when the rotation was large, the measured values would be smaller than the ground truth due to the local rotational slip between the sensor and the object (i.e., incipient slip). Thus, this method is more suitable under stable contact (less than 10 degrees in [12]). In contrast, our method could exclude the slip markers and utilize only the stick region for the calculation, thus improving in the accuracy. We describe the experimental results using the mean absolute rotational error (MARE) introduced in [13]. The results show that our method achieved a MARE of **0.17°±0.15°**.

### B. On-Line Measurement During Pivoting

The platform shown in Fig. 8(a) was built to conduct dynamic experiments. We mounted a self-made two-finger parallel gripper on a ROKAE XMate robot and installed two Tac3D tactile sensors on the end-effector. In this experiment, typical pivoting operations were performed by controlling the robot arm and the gripper to achieve the grasping and lifting of the in-hand object. We fixed a WT901BC angular sensor on the test objects to provide three-axis angles with an accuracy of 0.1° (as the ground truth). When the object was lifted, the Tac3D sensor and the angular sensor recorded data at 30 Hz.

A long-shaped object with a length of 250 mm was used to evaluate the performance by gripping and lifting tasks. As shown in Fig. 8(b), we staged three lifts after clamping the object, and recorded the ground truth values and the measured values in real time. Due to the camera noise, we applied a filtering operation with a window length of 5 and zero-drift compensation. The results exhibited three phases from stable contact to near-complete slip. In the first phase, the contact surface was almost fully sticking when the rotation was small. At the beginning of the incipient slip stage, the segment of the stick region was noisy since the slip amount was small, which caused obvious jitter in the results. As the slip accumulated during the plateau period, the results gradually approached flatness until the jitter reappeared in the next lifting period. Finally, the error amplified when the rotation angle increased further, and the state transitioned to macro slip. The tactile images of the three excerpted positions in Fig. 8(b) can visually explain the above discussion.

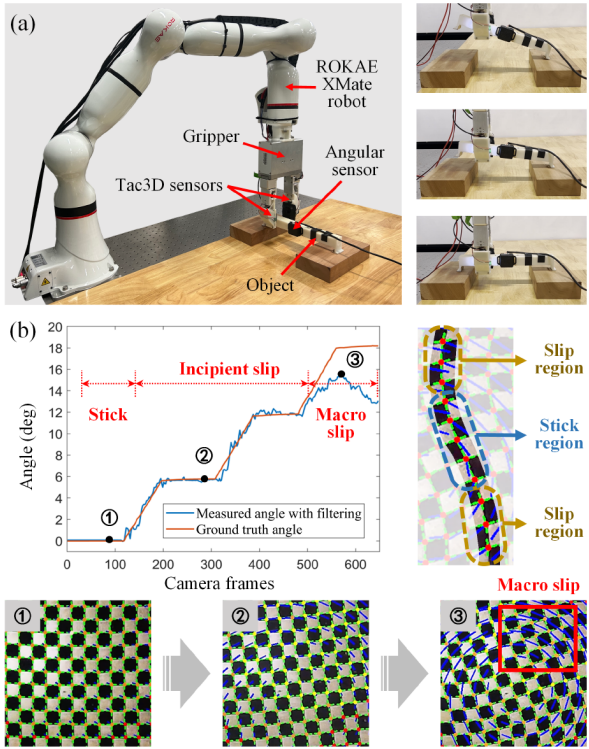

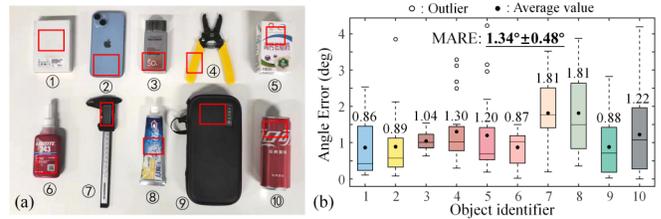

**Fig. 9.** (a) Experimental objects. (b) Rotation angle error between real-time measurement of Tac3D and ground truth value during the lifting phase.

**Fig. 8.** (a) Experimental setup for online rotation measurement. (b) Online measurements during lifting tasks using the long-shaped object shown in (a).

Ten household objects with different shapes, materials, and mass distributions were further selected [see Fig. 9(a)]. We conducted 20 to 25 pivoting experiments on each object, with each test consisting of a two-step operation of clamping and single lifting. At each lift to a random angle between 2 and 20 degrees, the deviation between the measured and the ground truth values was recorded. The statistical results of the measurement errors are shown in Fig. 9(b). The results indicate that the proposed method achieved a MARE of **1.34°±0.48°** in the adopted test object. According to the data provided by Castaño-Amorós *et al.* [13], the methods in [6], [12], [13], and [34] achieved MAREs of **3.96°±UNK**, **4.39°±0.18°**, **1.85°±0.96°** and **3.23°±1.69°**. Therefore, the proposed methods possessed advantages over state-of-the-art approaches even including learning-based techniques.

## VI. DISCUSSION

To summarize, this paper provides a theoretical analysis of the contact mechanics involved in pivoting, and achieves high accuracy and robustness to handle objects with various properties. The metric of line feature angles is sensitive to rotational slip while little affected by translation slip, and can effectively delineate the stick and slip regions on the pivoting contact surface. Compared to existing methods of the same type [12], [20], [33], we explore a non-empirical benchmark to distinguish between stick and slip markers, which provide more reliable and accurate results compared to manually selected strategies. The above characteristics make the method promising for benefiting diverse in-hand manipulation applications. For example, robots with pivot measurement capabilities can control the in-hand object poses to complete specific daily tasks (such as peg-on-hole and tool usage).

The proposed method still has limitations in handling objects under 3-d rotation or with soft structures. In Fig. 9(b), the results of object 7th and object 8th exhibit excessive errors compared to the other test objects. For the first scenario, the chosen vernier caliper was slender and prone to deflection during pivoting. Since the proposed method was based on a 2-d contact model, the related errors caused by theoretical simplification cannot be ruled out. In the second case, the selected toothpaste was a soft object. This issue highlights the deviation in the operation of simplifying Eq. (13) to Eq. (14). The above problems are common to existing approaches. For parallel grippers, the object softness has the greatest effect on estimation accuracy. Our simple test illustrated that when the object's softness was close to that of the elastomer, the measurement error reached about 70% (the pivoting angle was already difficult to define at this point). Besides, this approach is applicable for rotation measurement before macro slip. For tasks requiring controlled slip or rotations that exceed the deformation limits of the elastomer, we recommend tracking the rotation of the contact region contour.

## VII. CONCLUSION

In this paper, we introduce a generalized 2-d contact model considering incipient slip under in-hand object pivoting and propose a method for measuring the rotation angle based on line features. Experimental results show that the described approach could achieve a static MARE of 0.17°±0.15° and a dynamic MARE of 1.34°±0.48° without training, and was able to perform real-time perception in robotic pivoting tasks. Benefitting from the effective differentiation of stick/slip states through theoretical analysis, this approach achieves the state-of-the-art result even compared to related learning-based techniques. Future research will address its limitations in measuring 3-d object rotation and dealing with soft objects, and explore the possible enhancements brought by machine learning. We will also focus on designing suitable control strategies based on the accurate rotation information obtained to extend the proposed method to industrial applications.